\documentclass[runningheads]{llncs}
\usepackage{graphicx}
\usepackage{amsmath}
\usepackage{fontawesome}
\usepackage{todonotes}
\usepackage{svg}
\usepackage[hidelinks]{hyperref}
\usepackage{amsmath}
\usepackage{amssymb}
\usepackage{comment}
\usepackage{multirow}
\usepackage{orcidlink}
\usepackage[capitalize]{cleveref}

\newcommand{\norm}[1]{\left\lVert #1 \right\rVert}

\begin{document}

\title{Synthetic Augmentation for Anatomical Landmark Localization using DDPMs}
\titlerunning{Synthetic Augmentation for ALL using DDPMs}
% If the paper title is too long for the running head, you can set
% an abbreviated paper title here
%
\author{Arnela Hadzic \inst{1}$^\text{(\faEnvelope)}$\orcidlink{0009-0007-1476-3729}\and
Lea Bogensperger \inst{2}\and \\
Simon Johannes Joham \inst{1}\and 
Martin Urschler \inst{1}\orcidlink{0000-0001-5792-3971}
\thanks{This research was funded in whole or in part by the Austrian Science Fund (FWF) 10.55776/PAT1748423.}
}
\renewcommand\footnotemark{}
\authorrunning{A. Hadzic et al.}
% First names are abbreviated in the running head.
% If there are more than two authors, 'et al.' is used.
%
\institute{Institute for Medical Informatics, Statistics and Documentation,\\Medical University of Graz, Graz, Austria\\
\email{arnela.hadzic@medunigraz.at}\\ \and
Institute of Computer Graphics and Vision, Graz University of Technology,\\Graz, Austria}

\maketitle              % typeset the header of the contribution
\begin{abstract}
Deep learning techniques for anatomical landmark localization (ALL) have shown great success, but their reliance on large annotated datasets remains a problem due to the tedious and costly nature of medical data acquisition and annotation. While  traditional data augmentation, variational autoencoders (VAEs), and generative adversarial networks (GANs) have already been used to synthetically expand medical datasets, diffusion-based generative models have recently started to gain attention for their ability to generate high-quality synthetic images.
In this study, we explore the use of denoising diffusion probabilistic models (DDPMs) for generating medical images and their corresponding heatmaps of landmarks to enhance the training of a supervised deep learning model for ALL. Our novel approach involves a DDPM with a 2-channel input, incorporating both the original medical image and its heatmap of annotated landmarks. We also propose a novel way to assess the quality of the generated images using a Markov Random Field (MRF) model for landmark matching and a Statistical Shape Model (SSM) to check landmark plausibility, before we evaluate the DDPM-augmented dataset in the context of an ALL task involving hand X-Rays.
%The abstract should briefly summarize the contents of the paper in 15--250 words.
\end{abstract}
%
%
%

%%%%%%%%%%%%%%%%%%%%%%%%%%%%%%%%%%%%%%%%%%%%%%%%%%%%%%%%%%%%%%%%%%%%%%%%%%%%%%%%%
\section{Introduction}
Accurate localization of anatomical landmarks is essential in medical image analysis for various tasks, including segmentation~\cite{Heimann2009}, registration~\cite{Murphy2011}, and treatment planning~\cite{Wang2016}. State-of-the-art techniques for anatomical landmark localization (ALL) utilize deep learning models based on heatmap regression, demonstrating great success in automatically localizing landmarks in medical images~\cite{Tompson2014,Viriyasaranon2023}.
However, their reliance on large, annotated datasets remains a major challenge, as the process of annotating medical data is tedious and time-consuming, and acquiring medical images can be costly.

While some methods have successfully used limited medical datasets, techniques such as traditional data augmentation, variational autoencoders (VAEs), and generative adversarial networks (GANs) have been utilized to artificially expand medical datasets~\cite{Garcea2023,Pesteie2019,Frid2018,Chen2022}.
However, traditional data augmentation methods may not effectively cover the range of medical variability, as they can only generate variations of existing data. On the other hand, VAEs and GANs can learn the underlying data distribution and generate more diverse synthetic images. Nevertheless, GANs may encounter mode collapse issues, while VAEs are typically more stable to train but often suffer from low sample quality~\cite{Xiao2021}.

More recently, diffusion models have gained significant attention for their ability to generate high-quality images~\cite{Yang2023}. These models are a class of deep generative models used to model complex data distributions by iteratively denoising noise-perturbed samples. Their effectiveness has been demonstrated across a variety of computer vision tasks such as image generation~\cite{Dhariwal2021,Ho2020,Nichol2021}, image super-resolution~\cite{Yue2024,Zhu2023}, inpainting~\cite{Zhu2023}, segmentation~\cite{Amit2021,Bogensperger2023,Yu2023}, classification~\cite{Mukhopadhyay2023}, and anomaly detection~\cite{Wolleb2022Anomaly}, outperforming traditional VAEs and GANs in terms of image sample quality~\cite{Dhariwal2021,Moghadam2023,Kazerouni2023}. %Packhauser2023

In this study, we investigate the use of denoising diffusion probabilistic models (DDPMs) for generating synthetic medical hand images and their corresponding heatmaps of landmarks to facilitate the training of a supervised deep learning model for ALL. Up to our knowledge, we are the first to utilize a DDPM for this task, where we incorporate both the real hand images and their ground-truth heatmaps of landmarks as 2-channel inputs to a DDPM. We also propose a novel way to assess the quality of generated images using a Markov Random Field (MRF) model and a Statistical Shape Model (SSM) to fully automatically select the images that can be used as synthetic examples for training a supervised deep learning model.
In our ALL experiments, we utilize the Normalizing Flow-based Distribution Prior (NFDP)~\cite{Huang2024}, a {recent state-of-the-art ALL method, as a baseline. Our results demonstrate that incorporating DDPM synthetic images alongside real images for ALL yields improved results to those achieved with the baseline approach based solely on real images.

%%%%%%%%%%%%%%%%%%%%%%%%%%%%%%%%%%%%%%%%%%%%%%%%%%%%%%%%%%%%%%%%%%%%%%%%%%%%%%%%%
\section{Methods}
\subsection{Image Generation using DDPMs}
In order to generate pairs of synthetic medical hand images and their corresponding heatmaps of landmarks, we have designed a model based on a DDPM introduced in~\cite{Ho2020}. Our model learns to transform noisy samples into clean input samples by utilizing two processes: forward diffusion and reverse diffusion (\cref{fig:forward_reverse_diffusion}). In our approach, each input sample contains two channels - a 2D X-Ray image of a hand, and the corresponding heatmap image indicating the ground-truth landmarks on the hand. Once trained, the model is able to generate new pairs of medical hand images and their respective heatmaps of landmarks by sampling from the learned distribution.

\begin{figure}[h]
    \centering
    \includegraphics[width=\linewidth]{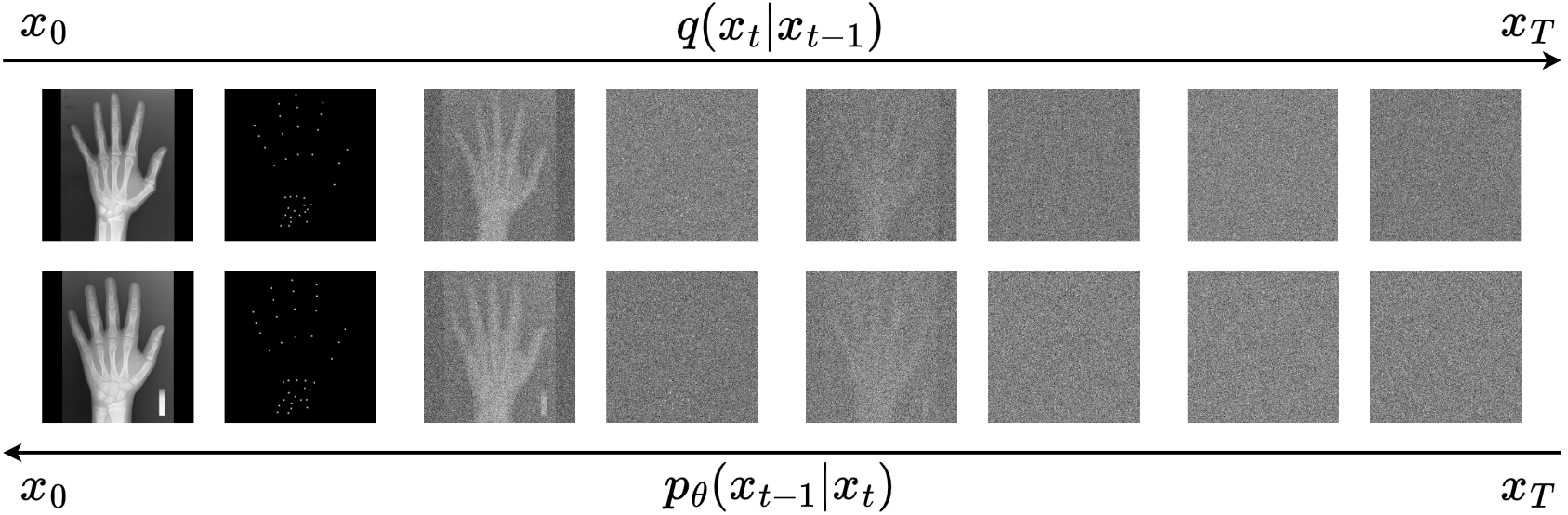}
    \caption{Illustration of the forward and reverse diffusion processes (best visible in the pdf version). The forward diffusion process $q(x_t|x_{t-1})$ introduces noise to a 2-channel input sample $x_0$ over T timesteps, while the reverse diffusion process $p_\theta(x_{t-1}|x_t)$ denoises the noisy sample $x_T$ to recover the original input sample $x_0$. } 
    \label{fig:forward_reverse_diffusion}
\end{figure}

\noindent In the forward diffusion process, a noisy sample $x_T$ is generated by gradually adding Gaussian noise to a given clean input $x_0$ according to a variance schedule $\beta_1, ..., \beta_T$ for a finite number of timesteps $T$. The process is formally defined as
\begin{equation}
    q(x_t|x_{t-1}):=\mathcal{N}(x_t;\sqrt{1-\beta_t}x_{t-1}, \beta_t I), 
\end{equation}
where the variances $\beta_t$ can be either learned through reparametrization or set as hyperparameters.

One key property of the forward diffusion process is that a noisy sample $x_t$, $t \in \{1, ..., T\}$, can be obtained directly from the initial input $x_0$:
\begin{equation}
    \begin{aligned}
    q(x_t|x_0):&=\mathcal{N}(x_t;\sqrt{\bar{\alpha}_t}x_{0}, (1-\bar{\alpha}_t) I)=\sqrt{\bar{\alpha}_t}x_{0} + \epsilon\sqrt{1-\bar{\alpha}_t}, \epsilon \sim \mathcal{N}(0, I), 
    \end{aligned}
    \label{eq:forward_process_property}
\end{equation}
where $\bar{\alpha}_t:=\prod_{s=1}^{t} \alpha_s$ and $\alpha_t:=1-\beta_t$.

The reverse diffusion process, defined by 
\begin{equation}
    p_\theta(x_{t-1}|x_t):=\mathcal{N}(x_{t-1};\mu_\theta(x_t, t), \Sigma_\theta(x_t, t)),
\end{equation}
involves the training of a neural network, to predict a mean $\mu_\theta(x_t,t)$ and a covariance matrix $\Sigma_\theta(x_t,t)$. The authors in~\cite{Ho2020} keep the covariance matrix fixed and perform a parametrization of $p_\theta(x_{t-1}|x_t)$, such that they can train a model $\epsilon_\theta(x_t,t)$ to directly predict noise $\epsilon$ from sample $x_t$ instead of predicting $\mu_\theta(x_t, t)$. The corresponding objective function is defined as follows:
\begin{equation}
    L(\theta):=\mathbb{E}_{t,x_0,\epsilon}\bigl[ \norm{\epsilon - \epsilon_\theta(\sqrt{\bar{\alpha}_t}x_0 + \sqrt{1-\bar{\alpha}_t}\epsilon, t)}^2 \bigl].
\end{equation}
To predict noise $\epsilon$, we utilize a U-Net architecture with residual blocks that is conditioned on the current timestep $t$ via sinusoidal position embedding.

\subsection{Quality Assessment of Generated Images}
\label{ssec:mrf_section}
For each image generated by our DDPM model, a corresponding heatmap of its landmarks is also generated. 
To automatically assess the quality of the generated image pairs, we evaluate them via landmark matching using an MRF model for labeling the landmarks in the generated heatmaps. After labeling, the plausibility of the landmarks is further analyzed using an SSM, where the labeled shapes are compared with the mean shape obtained from the training set.

\subsubsection{Landmark Matching with MRF}
A generated heatmap image contains blobs representing potential landmarks candidates. Inspired by~\cite{Donner2013}, the MRF's task is to label these candidates, such that the following energy function is maximized:
\begin{equation}
    f(c,e) = \sum_{\forall c_i \in C} U(c_i) + \sum_{\forall e_j \in E} B(e_j).
\end{equation}
Here, $U$ represents unary terms of nodes $c_i$ derived from a set of landmark candidates $C$, while $B$ represents binary terms of edges $e_j$ derived from a set of edges $E$. The pre-defined topology of our MRF model consists of L nodes connected by the edges $E$, and is derived from the training set (\cref{fig:mrf_topology} in \cref{app:mrf_topology}).

\begin{comment}
\begin{figure}[h]
    \centering
    \includegraphics[width=0.3\textwidth]{Figures/topology.png}
    \caption{MRF graph topology of a random training sample.}
    \label{fig:mrf_topology}
\end{figure}
\end{comment}

To compute the unary terms $U$, we apply a maximum filter on the generated heatmap image to identify L local maxima candidates for a landmark $l_i$, $i \in \{0, ..., \mathrm{L}-1\}$. These L candidates are then multiplied by values from individual 2D Gaussian matrices, each of them centered at the respective landmark's mean coordinate from the training set using a sigma value of 25 mm.
The resulting values measure the cost of assigning the candidates to the landmark $l_i$, thus defining the unary terms of our MRF model.

For the binary terms $B$, we model the distance distribution between landmark pairs using a t-distribution derived from the training set. 
The distance costs for a landmark pair $\langle l_i, l_j \rangle$, consisting of landmarks $l_i$ and $l_j$, are calculated as $d_{costs}=\mathrm{exp}(-\lvert d_{gt} - d \rvert)$. 
%\begin{equation}
%    d_{costs}=\mathrm{exp}(-\lvert d_{gt} - d \rvert), 
%\end{equation}
Here, $d_{gt}$ represents the distance between ground truth coordinates and $d$ represents the distance between predicted coordinates of the landmarks $l_i$ and $l_j$. The binary terms are then defined as $B=p(t_{distr}) + 2 d_{costs}$, 
%\begin{equation}
%    B=\mathrm{pdf}(t_{distr}) + 2 d_{costs}, 
%\end{equation}
where the first term represents a probability density function of the fitted t-distribution for the landmark pair $\langle l_i, l_j \rangle$, and the second term accounts for the previously defined distance costs.

We implement the MRF model as a factor graph and utilize the loopy belief propagation (LBP) optimizer to solve it. Once LBP converges, the landmarks are labeled for each synthetically generated heatmap image of the DDPM.

\subsubsection{Landmark Plausibility with SSM}
To check the plausibility of landmarks in generated heatmaps, we utilize an SSM~\cite{Cootes1995} defined as $x = x_{m} + Pb$.
%\begin{equation}
%    x = x_{m} + Pb.
%\end{equation}
Here, $x_m$ represents the mean shape obtained from aligned training shapes, $P$ is a matrix containing t eigenvectors of the covariance matrix, and $b$ is a t-dimensional vector of shape parameters. We then employ Random Sample Consensus to robustly match the mean shape configuration $x_m$ with the landmark configuration labeled by the MRF, using Euclidean distance as matching cost. This ensures that the majority of landmarks can still be matched even when some labeling errors occur. 

\subsubsection{Combining MRF and SSM}
After matching the shape configurations, we use following two constraints extracted from the training data statistics to determine MRF landmark configuration correctness: (i) no two landmarks can share the same coordinate, except for landmarks 2 and 3, as the ground truth coordinates of these landmarks are in very close proximity in the dataset; (ii) for every landmark in the wrist region, Euclidean distance between its labeled coordinate and the corresponding mean coordinate from the SSM must not exceed 16 mm. Only generated images whose corresponding heatmaps meet these two constraints are considered to be of high quality and thus accepted as synthetic data.

%%%%%%%%%%%%%%%%%%%%%%%%%%%%%%%%%%%%%%%%%%%%%%%%%%%%%%%%%%%%%%%%%%%%%%%%%%%%%%%%%
\section{Experimental Setup}
\setcounter{footnote}{0}
\subsection{Dataset}\label{section:dataset}
In this study, we use a publicly available dataset consisting of 895 2D X-Ray images of the left hand, with an average size of $1563\times2169$ pixels\footnote{\url{https://ipilab.usc.edu/research/baaweb}}. Each image in the dataset includes 37 ground truth landmarks provided by \cite{Payer2019}. Following the same normalization strategy as in \cite{Payer2019}, we assume a wrist width of 50 mm.

Our experimental setup consists of two main experiments: the FullDataset experiment and the ReducedDataset experiment. In the FullDataset experiment, we randomly split the dataset into a training set of 597 images and a test set of 298 images. For the ReducedDataset experiment, we randomly select 10\% images from the initial training set to form a new training set, while keeping the test set unchanged. In each experiment, we train a DDPM model on a respective training dataset and limit the number of generated images to 400, considering the long sampling time of diffusion models.

In order to evaluate the effectiveness of using synthetic data for ALL, we first train a supervised deep learning-based ALL network on the training sets from each experiment. We then evaluate its performance on the test set to establish a baseline. Next, we enhance the training sets by adding automatically selected DDPM synthetic images from each experiment and retrain the localization network. We then compare the performance of the network trained on both real and synthetic images to the baseline results. Evaluation metrics include the point-to-point error (PE) and the number of outliers $\mathrm{O}_r$ for radii of 2 mm, 4 mm, 10 mm, and 20 mm. Outliers for a radius $r$ are calculated as the total number of predicted landmarks that have a distance larger than $r$ from the ground truth.

Additionally, we compare the DDPM images with the images generated by a variational autoencoder (VAE) for both experiments. The quality of all images is evaluated using the MRF-SSM-based approach from Sect.~\ref{ssec:mrf_section}. For each experiment, we compute the mean acceptance rate defined as the ratio of the number of images selected by the MRF-SSM to the total number of generated images.

\subsection{Implementation Details}
To train the DDPM, we resize each image in the training set to $256\times256$ pixels and create a heatmap image of the same size based on the provided landmarks coordinates. Each heatmap is created with a sigma value of 1, corresponding to the standard deviation of the Gaussian function that defines the peak widths in the heatmap image. Both the hand images and their corresponding heatmaps are normalized to the range of [-1, 1] before being fed into the DDPM as a 2-channel input. The parameters for the forward diffusion process include 800 timesteps, a beta variance schedule ranging from $\beta_1=10^{-4}$ to $\beta_T=0.02$, and a batch size of 4. Similar to~\cite{Nichol2021}, our U-Net component in the reverse process consists of 3 levels and a bottleneck, each containing 3 residual blocks. Within each residual block, there are a group normalization layer and 2 convolutional layers, followed by SiLU activation. The residual blocks receive the time embedding through a linear layer, SiLU activation, and another linear layer. The network is trained over 10000 epochs with 32 input features, utilizing the Adam optimizer with a learning rate of 0.001 and the MSE loss function. The DDPM model is implemented in PyTorch and executed on an NVIDIA GeForce RTX 3090 Ti with 24 GB of GPU memory.

To compare the DDPM model with another deep generative model in terms of generated image quality, we trained a VAE to create $2\times256\times256$ hand and heatmap images in accordance with the DDPM setting. The VAE model is trained for 10000 epochs with a batch size of 32. Similar to the DDPM model, we use the Adam optimizer with a learning rate of 0.001 and 32 input features for the U-Net with residual blocks to enable a fair comparison. The latent dimension of the VAE is set to 64.

For the ALL baseline, we train the recent state-of-the-art Normalizing Flow-based Distribution Prior method~\cite{Huang2024}, which has a publicly available implementation\footnote{\url{https://github.com/jacksonhzx95/NFDP}}. We maintain all parameters consistent with their setup except for the number of epochs, which we increase to 2000. All images are resized to $512\times512$ pixels for both training and testing.

%%%%%%%%%%%%%%%%%%%%%%%%%%%%%%%%%%%%%%%%%%%%%%%%%%%%%%%%%%%%%%%%%%%%%%%%%%%%%%%%%
\section{Results and Discussion} 
In contrast to previous studies that have not reported how they assess the quality of synthetically generated images~\cite{Yu2023} or relied on metrics like Fréchet Inception Distance (FID) score, which may not be suitable for medical images due to being based on a network trained on natural images~\cite{Ye2023}, our study utilized an automated evaluation method based on MRF and SSM to assess the realism of the generated images. This approach allowed us to fully automatically select image/heatmap pairs with realistic features.
To quantitatively compare the performance of the DDPM and VAE models, we sampled 100 images in ten independent runs using each model. As shown in Table~\ref{table:ddpm_vae_comparison}, in the FullDataset setup, the DDPM model achieved a mean acceptance rate of 40\% ($\pm$ 3\% SD), while the VAE model had a mean acceptance rate of only 3\% ($\pm$ 3\% SD). When the training data was limited to 60 images (10\% of the initial training set), the DDPM model achieved a mean acceptance rate of 26\% ($\pm$5\% SD). Differently, the VAE model was unable to generate realistic images from only 60 training samples. \cref{fig:generated_samples} visualizes the qualitative results of the DDPM, showcasing examples of images that were automatically selected by the MRF-SSM, as well as examples of images that were automatically rejected, since they did not meet the criteria for correctness. Additional examples of DDPM and VAE generated images can be found in \cref{fig:ddpm_full,fig:ddpm_10_percent,fig:vae_full} of \cref{app:additional_results}.

\begin{figure}[t]
    \centering
    \includegraphics[scale=0.32]{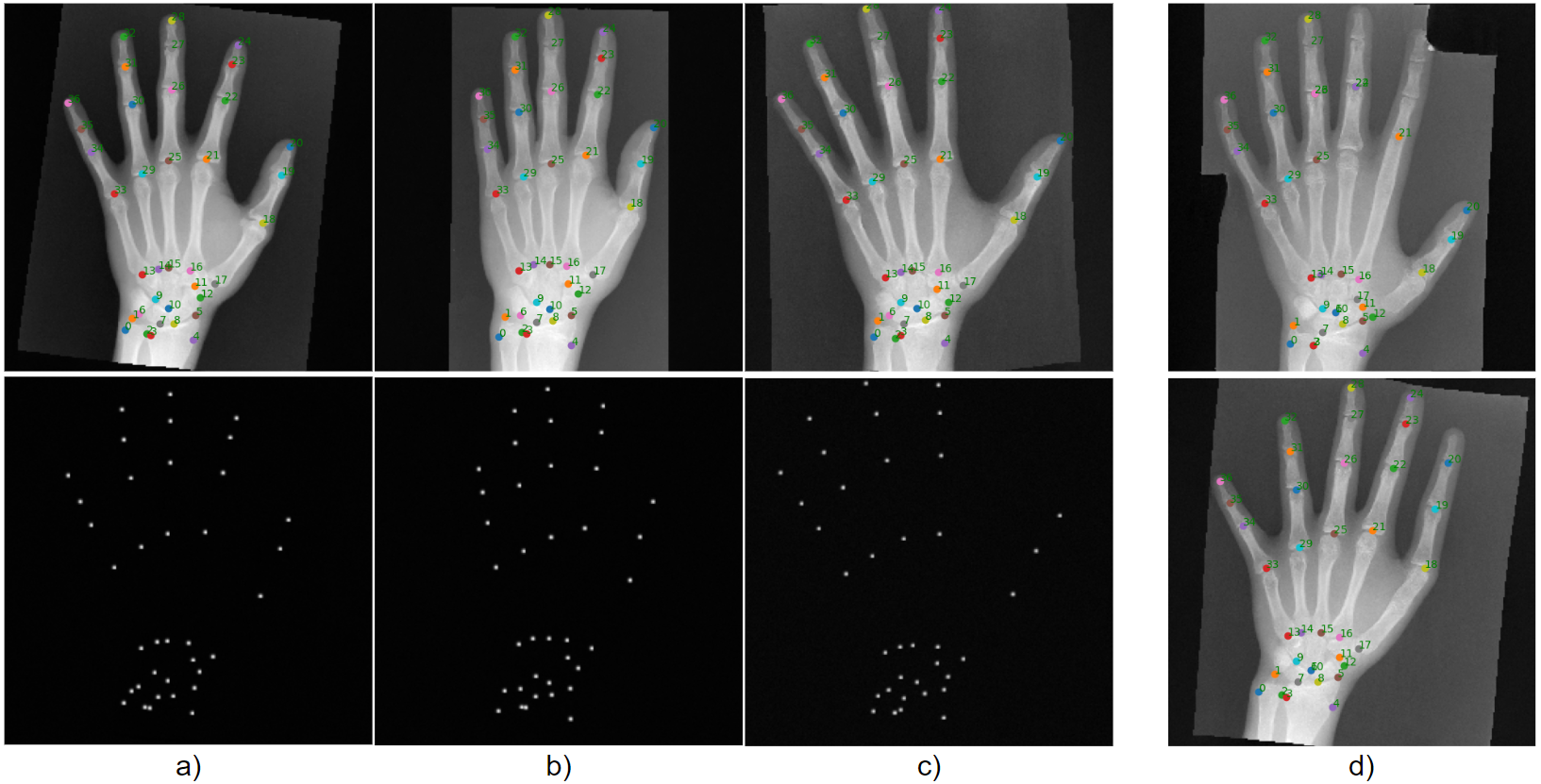}
    \caption{Qualitative results of DDPM samples with MRF labeling (colored points). Samples a) and b) were automatically selected in the FullDataset experiment, c) was selected in the ReducedDataset experiment, while images like in d) were automatically rejected via our proposed assessment strategy due to abnormalities such as six fingers or a very long thumb.
    }
    \label{fig:generated_samples}
\end{figure}

\begin{table}[h]
    \caption{DDPM and VAE comparison.}
    \centering
    \begin{tabular}{l|l|l|l|l}
    \hline
    Metric & DDPM 100\% $\uparrow$ & DDPM 10\% $\uparrow$ & VAE 100\% $\uparrow$ & VAE 10\% $\uparrow$ \\ \hline \hline
    mean acceptance rate & 40\% ($\pm$3\% SD) & 26\% ($\pm$5\% SD) & 3\% ($\pm$3\% SD) & 0\% \\ \hline
    \end{tabular}
    \label{table:ddpm_vae_comparison}
\end{table}

\noindent For the ALL task, we conducted experiments training the NFDP on a combination of real and synthetic images and compared the results with the NFDP trained solely on real images. The results outlined in Table~\ref{table:results} show that incorporating DDPM synthetic images into the training yielded slightly better results than the NFDP trained without synthetic data. In contrast, the NFDP trained on real and VAE synthetic images performed worse than the NFDP trained only on real images. This suggests that even with our proposed quality assessment strategy, incorporating additional synthetic images from VAE cause performance degradation, likely due to the poor image quality of VAE.
In the ReducedDataset experiment, the point-to-point error (PE) and the number of outliers for radii of 2 mm and 4 mm did not show improvement compared to training solely on real images, which may be attributed to the low number of real training images. However, incorporating DDPM synthetic data into the training of the ALL network did result in a decrease in the number of outliers for larger radii of 10 mm and 20 mm. The NFDP evaluation on real and VAE synthetic images was not possible in the ReducedDataset scenario due to the 0\% mean acceptance rate of the VAE generated images. Overall, this suggests the potential of synthetically generated images using DDPMs for improving robustness in ALL, which is crucial for downstream applications like segmentation that rely on accurately identified landmarks for subsequent processing~\cite{Hadzic2023}.

\begin{figure}[t]
    \centering
    \includegraphics[scale=0.36]{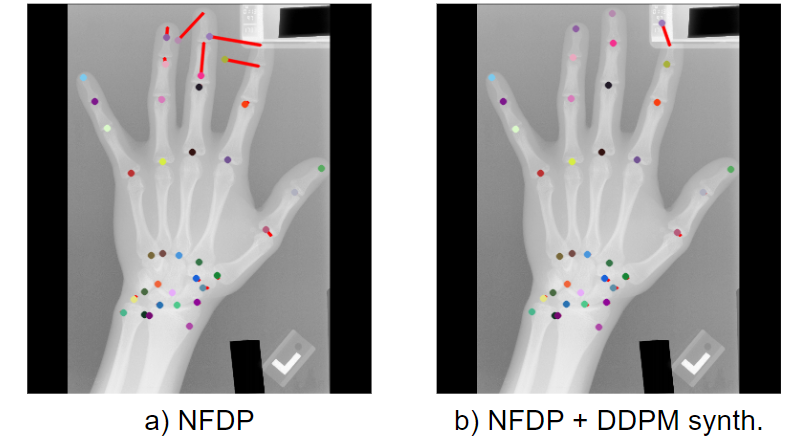}
    \caption{Example of occluded fingertip, where synthetic DDPM augmentation enables recovery from localization errors. Predicted landmarks are represented by colored points, with red lines indicating their distance to the ground truth landmarks.}
    \label{fig:occlusion_outliers}
\end{figure}

\begin{table*}[h]
    \caption{Results of the NFDP network trained with and without synthetic images.}
    \centering
    \begin{tabular}{l|l|l|l|l|l}
        \hline
        \multirow{2}{*}{NFDP} & PE in mm $\downarrow$ & \multirow{2}{*} {$\text{O}_{r=2\mathrm{mm}}$ $\downarrow$}  & \multirow{2}{*}{$\text{O}_{r=4\mathrm{mm}}$ $\downarrow$} & \multirow{2}{*}{$\text{O}_{r=10\mathrm{mm}}$ $\downarrow$} & \multirow{2}{*}{$\text{O}_{r=20\mathrm{mm}}$ $\downarrow$}\\
            & Mean $\pm$ SD & & & & \\
        \hline
        \hline
        100\% \ (real) & 0.76 \ $\pm$ 0.79 & 702 & 92 & \textbf{1} & \textbf{0}  \\
        100\% \ (real + VAE) & 0.77 \ $\pm$ 0.79 & 711 & 91 & \textbf{1} & \textbf{0} \\
        100\% \  (real + DDPM) & \textbf{0.75} $\pm$ \textbf{0.77} & \textbf{697} & \textbf{84} & \textbf{1} & \textbf{0} \\
        \hline
        10\% \ \ (real) & \textbf{0.82} $\pm$ 0.84  & \textbf{776} & \textbf{91} & 3 & 1 \\
        10\% \ \ (real + VAE) & / & / & / & / & / \\
        10\% \ \ (real + DDPM) & 0.84 \ $\pm$ \textbf{0.81} & 840 & 101 & \textbf{1} & \textbf{0} \\ \hline
    \end{tabular}
    \label{table:results}
\end{table*}

\noindent The reduction of large outliers using DDPM synthetic images can also be demonstrated in an occlusion scenario. \cref{fig:occlusion_outliers} shows a hand image from the 2D X-Ray dataset with an occluded landmark. This particular image was not included in the training set for the ReducedDataset experiment, allowing us to study the NFDP in this setup.
\cref{fig:occlusion_outliers}a) shows predictions of the NFDP trained solely on real images, whereas \cref{fig:occlusion_outliers}b) shows predictions of the NFDP trained on a combination of real and DDPM synthetic images. Despite the occlusion in the image, our method only had one landmark with a distance larger than 10 mm from the ground truth, as shown in \cref{fig:occlusion_outliers}b). On the other hand, \cref{fig:occlusion_outliers}a) shows four landmarks with a distance larger than 20 mm. This suggests that utilizing DDPM synthetic images may enhance the generalization of the ALL network.

Overall, we assume that the promising performance improvement comes from additional variation of shape introduced by our DDPM generated images, which seems to be beneficial for the robustness of the supervised ALL method, especially in occlusion scenarios. In future work, we plan to conduct a comparison with GANs and explore how to accelerate the sampling process in DDPMs.

%%%%%%%%%%%%%%%%%%%%%%%%%%%%%%%%%%%%%%%%%%%%%%%%%%%%%%%%%%%%%%%%%%%%%%%%%%%%%%%%%
\section{Conclusion}
In this study, we introduced a DDPM to generate pairs of medical hand images and their corresponding heatmaps of landmarks for synthetically augmenting anatomical landmark localization. To assess the quality of synthesized images, we utilized an MRF model and an SSM to eliminate unrealistic images. Compared to a baseline approach for landmark localization, the experimental results introducing synthetic images during training showed improvements in terms of reducing large outliers, which is essential in downstream tasks such as segmentation. 
This suggests the potential of utilizing DDPMs to enhance the training data of supervised networks in the context of anatomical landmark localization.

\begin{appendix}
\section{MRF Topology}\label{app:mrf_topology}
\begin{figure}[h]
    \centering
    \includegraphics[scale=0.5]{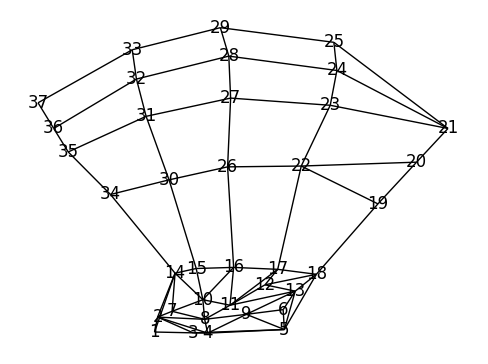}
    \caption{MRF topology of a random training sample.}
    \label{fig:mrf_topology}
\end{figure}

\clearpage

\section{Additional Qualitative Examples}\label{app:additional_results}
\begin{figure}[h!]
    \centering
    \includegraphics[width=\textwidth]{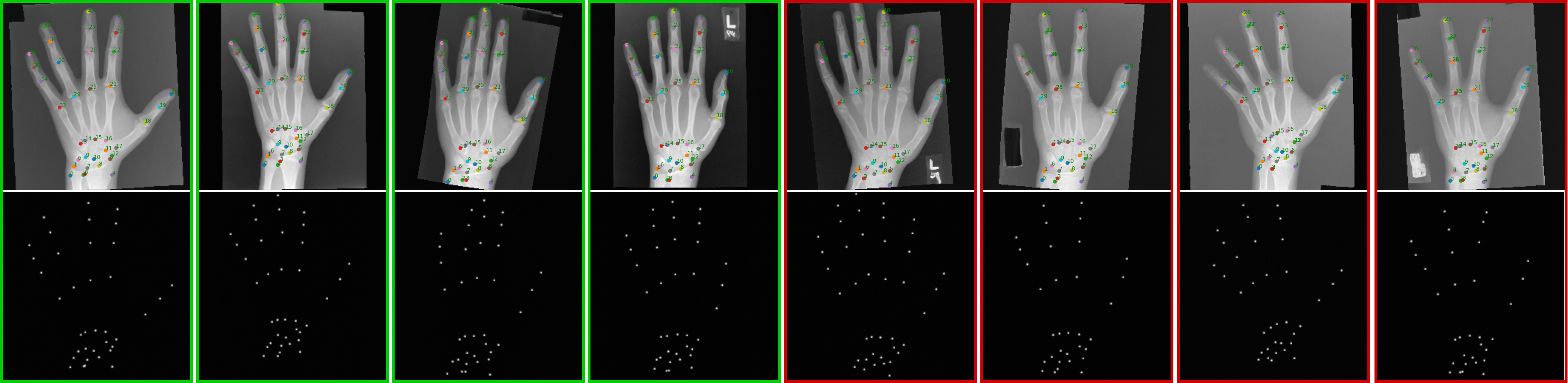}
    \caption{Image/heatmap pairs generated by the DDPM in the FullDataset scenario (green frame - automatically accepted samples, red frame - automatically rejected samples).}
    \label{fig:ddpm_full}
\end{figure}
\begin{figure}[h!]
    \centering
    \includegraphics[width=\textwidth]{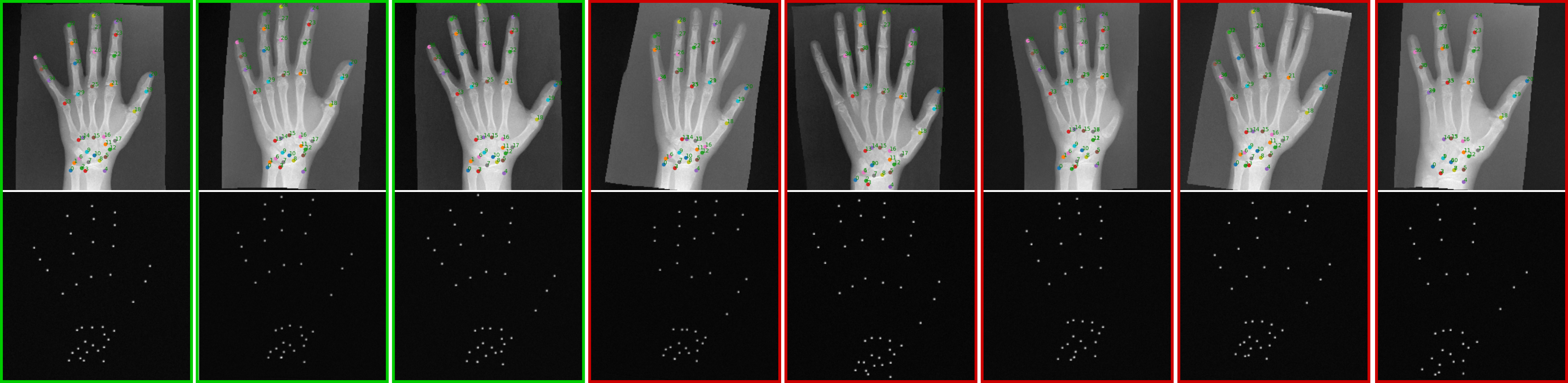}
    \caption{Image/heatmap pairs generated by the DDPM in the ReducedDataset scenario (green frame - automatically accepted samples, red frame - automatically rejected samples).}
    \label{fig:ddpm_10_percent}
\end{figure}
\begin{figure}[h!]
    \centering
    \includegraphics[width=\textwidth]{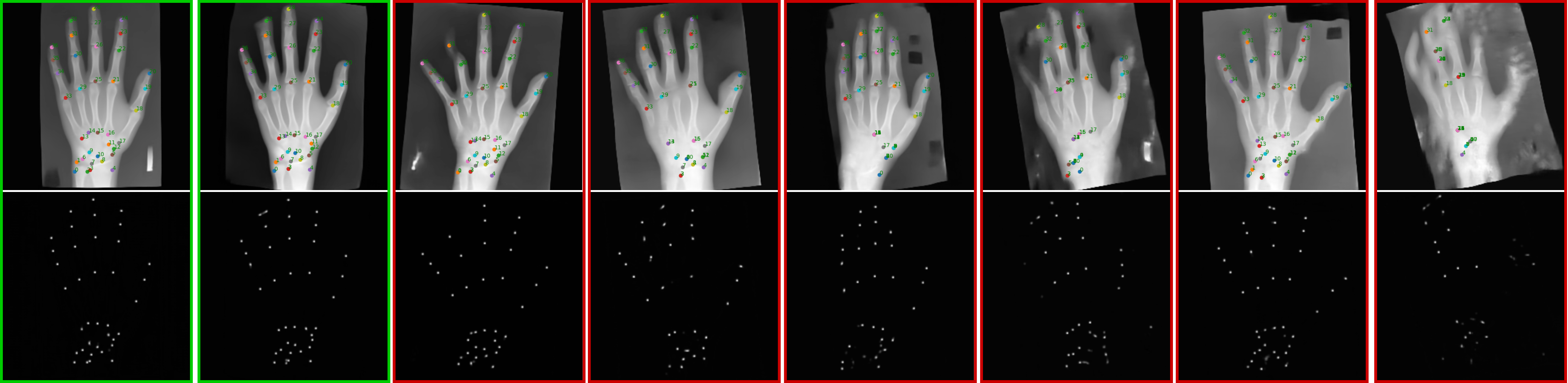}
    \caption{Image/heatmap pairs generated by the VAE in the FullDataset scenario (green frame - automatically accepted samples, red frame - automatically rejected samples).}
    \label{fig:vae_full}
\end{figure}
\end{appendix}
\clearpage

%
% ---- Bibliography ----
%
% BibTeX users should specify bibliography style 'splncs04'.
% References will then be sorted and formatted in the correct style.
%
\bibliographystyle{splncs04}
\bibliography{samplepaper}
%
%\begin{thebibliography}{8}
%\bibitem{ref_article1}
%Author, F.: Article title. Journal \textbf{2}(5), 99--110 (2016)

%\bibitem{ref_lncs1}
%Author, F., Author, S.: Title of a proceedings paper. In: Editor,
%F., Editor, S. (eds.) CONFERENCE 2016, LNCS, vol. 9999, pp. 1--13.
%Springer, Heidelberg (2016). \doi{10.10007/1234567890}

%\bibitem{ref_book1}
%Author, F., Author, S., Author, T.: Book title. 2nd edn. Publisher,
%Location (1999)

%\bibitem{ref_proc1}
%Author, A.-B.: Contribution title. In: 9th International Proceedings
%on Proceedings, pp. 1--2. Publisher, Location (2010)

%\bibitem{ref_url1}
%LNCS Homepage, \url{http://www.springer.com/lncs}. Last accessed 4
%Oct 2017
%\end{thebibliography}
\end{document}